%% file: main.tex
\documentclass{article}

\PassOptionsToPackage{numbers}{natbib}
\usepackage[preprint]{neurips_2021}

\usepackage[utf8]{inputenc} 
\usepackage[T1]{fontenc}    
\usepackage{hyperref}       
\usepackage{url}            
\usepackage{booktabs}       
\usepackage{amsfonts}       
\usepackage{nicefrac}       
\usepackage{microtype}      
\usepackage{xcolor}         
\usepackage{tikz}
\usepackage{pgfplots}     
\pgfplotsset{width=9cm,compat=1.3}
\definecolor{red1}{rgb}{0.6,0.05,0.1}
\definecolor{red2}{rgb}{0.8,0.4,0.35}
\definecolor{red3}{rgb}{1,0.1,0.05}
\definecolor{blue1}{rgb}{0.3,0.3,0.6}
\definecolor{blue2}{rgb}{0.35,0.3,1}
\definecolor{blue3}{rgb}{0.08,0,1}
\definecolor{yellow}{rgb}{1, 0.95, 0.18}

\usepackage{multirow}

\usepackage{amsmath}
\DeclareMathOperator*{\argmax}{arg\,max}

\title{Why Out-of-distribution Detection in CNNs Does Not Like Mahalanobis - and What to Use Instead}

\author{%
 Kamil Szyc \\
    Wrocław University of Science and Technology\\
    11/17 Z. Janiszewskiego Str. \\
    50-372 Wrocław, Poland \\
  \texttt{kamil.szyc@pwr.edu.pl}\\
  \AND
 Tomasz Walkowiak \\
    Wrocław University of Science and Technology\\
    11/17 Z. Janiszewskiego Str. \\
    50-372 Wrocław, Poland \\
  \texttt{tomasz.walkowiak@pwr.edu.pl}\\
  \AND
  Henryk Maciejewski\\
    Wrocław University of Science and Technology\\
    11/17 Z. Janiszewskiego Str. \\
    50-372 Wrocław, Poland \\
  \texttt{henryk.maciejewski@pwr.edu.pl}\\
}

\begin{document}

\maketitle

\begin{abstract}
Convolutional neural networks applied for real-world classification tasks need to recognize inputs that are far or out-of-distribution (OoD) with respect to the known or training data. To achieve this, many methods estimate class-conditional posterior probabilities and use confidence scores obtained from the posterior distributions. Recent works propose to use multivariate Gaussian distributions as models of posterior distributions at different layers of the CNN (i.e., for low- and upper-level features), which leads to the confidence scores based on the Mahalanobis distance. However, this procedure involves estimating probability density in high dimensional data using the insufficient number of observations (e.g. the dimensionality of features at the last two layers in the ResNet-101 model are 2048 and 1024, with ca. 1000 observations per class used to estimate density). In this work, we want to address this problem. We show that in many OoD studies in high-dimensional data, LOF-based (Local Outlierness-Factor) methods outperform the parametric, Mahalanobis distance-based methods. This motivates us to propose the nonparametric, LOF-based method of generating the confidence scores for CNNs. We performed several feasibility studies involving ResNet-101 and EffcientNet-B3, based on CIFAR-10 and ImageNet (as known data), and CIFAR-100, SVHN, ImageNet2010, Places365, or ImageNet-O (as outliers). We demonstrated that nonparametric LOF-based confidence estimation can improve current Mahalanobis-based SOTA or obtain similar performance in a simpler way.
\footnote{All results are fully reproducible, and the code is available on GitHub upon request.}

\end{abstract}

\input{sections/introduction.tex}

\label{section:method}
\input{sections/method.tex}\texttt{}

\label{section:analysis}
\input{sections/analysis.tex}

\input{sections/conclusion.tex}

\bibliographystyle{plainnat}
\bibliography{bibliography}

\end{document}

%% file: sections/introduction.tex
\section{Introduction}

\begin{figure}
  \centering
  \label{plot:idea}
  \includegraphics[width=0.95\textwidth]{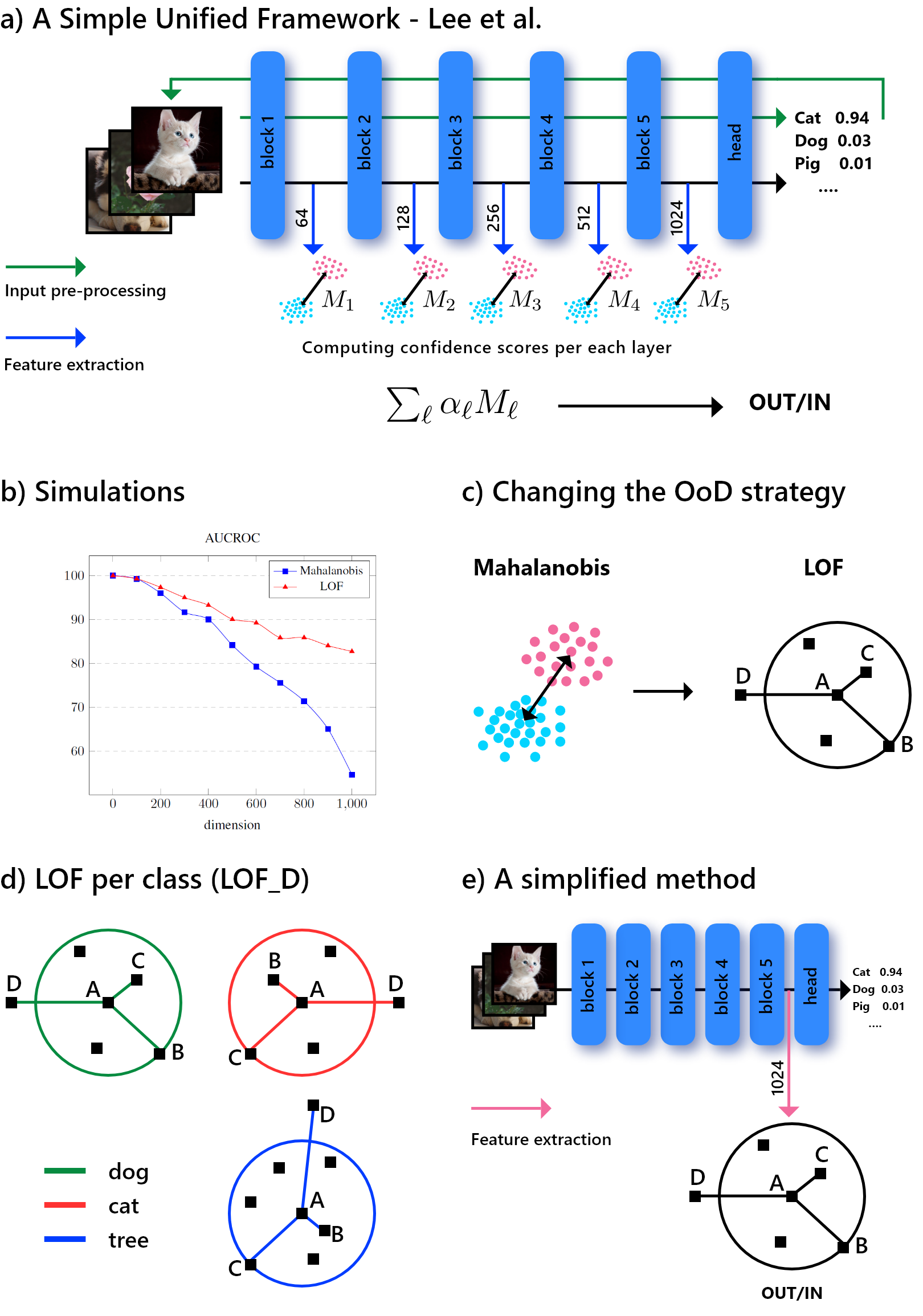}
  \caption{The idea of the proposed method. \textbf{a)} Method by \citet{lee2018simple}, which uses the input pre-processing \cite{liang2017enhancing}, with Mahalanobis distance-based confidence score $M_{\ell}$ from feature ensemble. For hyperparameter tuning, some OoD examples are needed. \textbf{b)} We show that non-parametric LOF outperforms Mahalanobis as an OoD detector in high-dimensional data. \textbf{c)} We propose to compute confidence scores using LOF in place of Mahalanobis distance. \textbf{d)} We propose a modified version of LOF - LOF\_D, which achieves good results for some CNN models. \textbf{e)} Simplified version of our method: OoD detection only using the last layer of DNN - successful on many datasets, hyperparameter free, and do not use any OoD data.
}
\end{figure}
Deep neural networks deployed for real-world recognition tasks need to recognize inputs far from the known or training data. However, despite high classification accuracy on the benchmark datasets (e.g., top-1 accuracy on the ImageNet equal 84.4\% for EfficientNet-B7 \citep{tan2019efficientnet}, and 90.2\% for one of the latest models by \citet{pham2020meta}), deep networks are vulnerable to out-of-distribution (OoD) or adversarial inputs that are easily recognized by humans \citep{chakraborty2018adversarial}, \citep{hendrycks2019natural}, \citep{zhou2019humans}, \citep{nguyen2015deep}.  Recognition of out of distribution examples is essential in safety-critical applications or incremental-learning systems \citep{amodei2016concrete}, \citep{feng2018towards}, \citep{sharif2016accessorize},  \citep{evtimov2017robust}. 

Many current approaches recognize inputs that are unlike the known data by estimating class-conditional posterior probabilities. Posterior distributions are then used to obtain confidence scores of prediction. Some authors propose to use multivariate Gaussian distributions as models of class-conditional distributions \citep{bendale2015towards}, \citep{lee2018simple}, \citep{sehwag2021ssd}, which leads to estimating the uncertainty of prediction using Mahalanobis distance. Alternative methods attempt to recognize out of distribution examples by quantifying outlierness $d(x, X_i)$ of input $x$ with regard to the known training data $X_i$ related to the class $c_i$ maximizing the posterior probability $p(c|x)$, ie. $c_i = \argmax_{c \in C} P(c | x)$. The score $d$ is estimated as distance, density, or some measure of outlierness, such as the Local Outlierness Factor (LOF) \citep{Breunig-2000}. A recent comprehensive survey of these approaches is given in \citep{geng2020recent}. 

Many recent methods of detection of OoD inputs in deep neural networks are based on confidence scores from posterior distributions. The baseline method by \citet{hendrycks2016baseline} is based on the maximum value of posterior distribution obtained as the softmax score. The ODIN method by \citet{liang2017enhancing} uses input preprocessing and temperature scaling in the softmax function to increase separability between in- and out-of-distribution samples. The unified framework proposed by \citet{lee2018simple} uses the Mahalanobis distance-based score aggregated from many layers of the DNN (ie. for lower and upper-level features). While these methods use standard representations learned to discriminate within in-distribution classes, some recent works e.g., \citep{sehwag2021ssd}, \citep{tack2020csi} propose to use contrastive learning-based representations which aim to discriminate in- and OoD samples.   

Some authors propose to model posterior distributions using multivariate Gaussian distribution, which leads to confidence scores based on Mahalanobis distance.
However, these procedures require estimating probability densities in high dimensional representations generated by the DNN (e.g. the dimensionality of features at the last layer of the ResNet-101\cite{he2016deep} is 2048, and of EfficientNet-B3 is 1536), commonly using ca. 1000-2000 observations per class. This problem motivated this work. We argue that the quality of such parametric density estimates is questionable and that more reliable confidence scores for OoD detection can be proposed using the nonparametric LOF method.

Our main contributions are the following. We show that the LOF-based method outperforms the Mahalanobis distance-based method in the task of OoD detection in high-dimensional data (section \ref{section:Maha-vs-LOF}). Motivated by this, we propose a simple method for detecting OoD inputs, inspired by the unified framework by \citet{lee2018simple}, with confidence scores obtained using the LOF instead of the Mahalanobis distance. Similar to \citep{lee2018simple}, our method uses standard representations generated by any modern neural network architecture and does not require re-training the model. The idea of the proposed method is summarized in Figure \ref{plot:idea}.
We performed several feasibility studies using ResNet-101 and EffcientNet-B3, trained on CIFAR-10 and ImageNet, with CIFAR-100, SVHN, ImageNet2010,
Places365, or ImageNet-O used as outliers. We demonstrate that nonparametric, LOF-based OoD detection can improve Mahalonobis-based results, or on some datasets obtain similar performance in a simpler way, ie. with fewer (or no) hyperparameters, and with no OoD samples needed for hyperparameter tuning.

%% file: sections/method.tex
\section{Estimation of confidence scores using LOF-based outlierness measure}
\subsection{Mahalanobis distance-based vs nonparametric outlierness factor-based OoD in high dimensional data}
\label{section:Maha-vs-LOF}

Since the estimation of density in high dimensional data is generally considered unattractive \citep{hastie2009elements}, we want to empirically show confidence scores obtained from density estimates lead to the limited performance of OoD detection. On the other hand, outlier or out-of-distribution detection in high dimensional data can be performed reasonably well using scores obtained from the Local Outlierness Factor method \citet{Breunig-2000}.

We performed a simple simulation study in which we compared the performance of OoD detection based on confidence scores obtained using the Mahalanobis distance vs confidence scores obtained with the LOF algorithm. As in-distribution (known) data we generated two clusters from the MVN distribution in $d$ dimensions, with the mean at $[0]_d$ and $[-1]_d$ and uncorrelated variables with variance 1. As OoD we used a cluster with mean at $[\frac{r}{\sqrt{d}}]_d$, uncorrelated, with variance 1. Confidence scores were calculated as the Mahalanobis distance between a test sample and the closest class-conditional Gaussian distribution, which can be interpreted as the log of the probability density of the test sample. In the alternative approach, confidence scores were obtained as local outlierness factors (LOF) calculated for test samples with respect to the closest cluster of known data (see section \ref{section:OoD-with-LOF} for technical details of LOF).

\input{figures/LOFvsMah}

In Figure \ref{plot:LOFvsMahalSim}, we summarize performance of Mahalanobis distance-based and LOF-based OoD detectors as a function of dimensionality $d$. 
We observe that for $d=1000$ with 1000 training points per class, the Mahalanobis procedure no longer detects outliers (AUCROC $\approx$ 50\%), while LOF is more reliable (AUCROC > 80\%). The same effect is observed if the probability density of the known data is modeled using the Weibull distribution fitted to data by the EVM \citet{rudd2017extreme} algorithm (AUCROC $\approx$ 50\% for $d=1000$).   

We conclude that in high dimensional data ($d=1000 \dots 2000$) OoD detection can be done more reliably is the LOF-based score is used in place of the scores obtained from parametric density estimates, such as Mahalanobis distance-based.

\subsection{LOF-based outlierness measures}

\label{section:OoD-with-LOF}
The Local Outlier Factor \cite{Breunig-2000} (LOF) is based on an analysis of the local density of points. It works by calculating the so-called local reachability density $LRD_k(x,X_i)$ of input $x$ with regard to the known dataset $X_i$. $LRD$ is defined as an inverse of an average reachability distance between a given point, its $k$-neighbors, and their neighbors (for details refer to \cite{Breunig-2000}). 
$K$-neighbors ($N_k(x,X_i)$) includes a set of points that lie in the circle of radius $k$-distance, where $k$-distance is the distance between the point, and it’s 
the farthest $k^{th}$ nearest neighbor ($||N_k(x,X_i)||>=k$).
The local outlier factor (LOF) is formally defined as the ratio of the average $LRD$ of the $k$-neighbors  of the point $x$ to the $LRD$ of the point. 
\begin{equation}
   d_{LOF}(x, X_i) =  \frac{\sum_{y\in{N_k(x,X_i)}}{LRD_k(y,X_i)}}{||N_k(x,X_i)||LRD_k(x,X_i)}
\label{equation:LOF}
\end{equation}

Intuitively, if the point is an inlier, the ratio of average $LRD$ of neighbors is similar to the $LRD$ of the point. So the LOF is around 1. For outliers it should be above 1 since the density of an outlier is smaller than its neighbor densities.

Original LOF\cite{Breunig-2000} was used for outlier detection inside a data set but it can be easily extended to out-of-distribution recognition. We can calculate local reachability densities for all training data and treat them and all training data as the model. During outlierness quantification of input $x$, we have to find its nearest neighbors among all training data, calculate the local reachability density for $x$ and divide the average $LRD$ of neighbors by $x$ $LRD$. Original LOF\cite{Breunig-2000} is based on Euclidean distance and we will use this distance in all experiments marked as LOF.
Moreover, we modified the original LOF by using the cosine metric and building a separate LOF model for each in-distribution class (see Figure \ref{plot:idea} \textbf{d)}). For OoD detection/uncertainty quantification, we use the model LOF that corresponds to the closest class identified by the CNN. We refer to this approach as LOF\_D. Because the LOF\_D method builds separate models for each class, the time of determining the confidence score is at least $n$ times shorter than in the case of the LOF method, where $n$ is the number of classes.
The use of LOF or LOF\_D requires specifying the number of analyzed neighbors $k$. The impact of this value on the effectiveness of OoD detection for relatively large data sets that are analyzed in this work is negligible. Therefore, in all reported experiments  $k = 20$ is assumed (which is the default value in the LOF implementation \footnote{\url{https://scikit-learn.org/stable/modules/generated/sklearn.neighbors.LocalOutlierFactor.html}}).

\subsection{OoD in CNN using LOF-based outlierness measure}
\label{section:SUF-with-LOF}

The Simple Unified Framework by \citet{lee2018simple} is based on three techniques to detect OoD in CNNs (see Figure \ref{plot:idea} \textbf{a)}), i.e. the Mahalanobis distance-based confidence scores, ensembles of scores from each layer of the CNN, and the input pre-processing by adding small perturbations (ODIN technique proposed by \citet{liang2017enhancing}). We propose to replace the Mahalanobis distance-based confidence scores by LOF-based outlierness scores - Eq. \ref{equation:LOF}, and omit the ODIN-like input preprocessing (see Figure \ref{plot:idea} \textbf{c)}). The first modification is inspired by the simulation experiments described in Section \ref{section:Maha-vs-LOF} (see Figure \ref{plot:idea} \textbf{b)}). The second modification is motivated by the fact that input perturbations are computationally complex and require parameter-tuning (the magnitude of perturbation). In our experiments, we found that the ODIN-like input perturbations have negligible (or even negative) impact on separating the in- and out-of-distribution samples for LOF-based methods. For example, for experiments with CIFAR-10 (see Section \ref{section:experiment_organization}) perturbation equal zero was always selected as the best on validation data (we used the same fine-tuning method as proposed in \citet{lee2018simple}).
In contrast to this observation, the Mahalanobis distance-based technique for OoD detection by \citet{lee2018simple} is sensitive to the correct choice of the input perturbation strength. Apparently, the Mahalanobis distance-based method requires better separation of the in- and out-of-distribution samples compared with the non-parametric LOF outlierness detector. 

In the feasibility studies in Section \ref{section:sanalysis}, we also report the performance of the simplified version of the proposed method (see Figure \ref{plot:idea} \textbf{e)}), with the estimation of the outlierness using only the last (perultimate) layer of the deep model. We found that for some simple OoD detection problems, the simplified method performs reasonably well, which yields a hyperparameter-free OoD detection method.

%% file: figures/LOFvsMah.tex
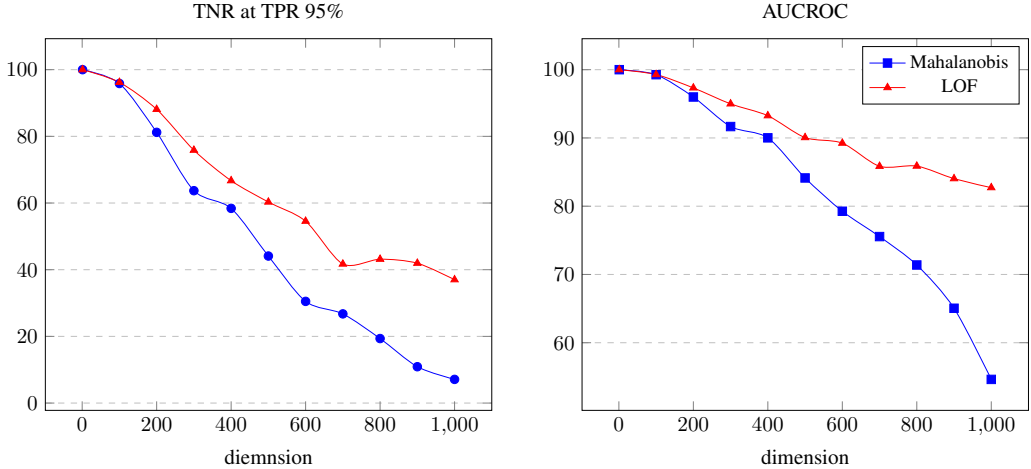
\begin{figure}[htb]
\centering
\begin{tikzpicture}[xscale=0.80, yscale=0.8]
    \begin{axis}[
    name=ax1,
    title={TNR at TPR 95\%},
    xlabel={diemnsion},
    ylabel={},
    ymajorgrids=true,
    grid style=dashed,
]

\addlegendentry{Mahalanobis}
\addplot[smooth,mark=*,blue,solid]  
    coordinates {
    (1,100)
(100,95.85)
(200,81.2)
(300,63.7)
(400,58.4)
(500,44.1)
(600,30.5)
(700,26.75)
(800,19.35)
(900,10.9)
(1000,7.1)
    };
\addlegendentry{LOF}
\addplot[smooth,color=red,mark=triangle*,solid] 
    coordinates {
   (1,100)
(100,96.2)
(200,88.1)
(300,75.85)
(400,66.7)
(500,60.3)
(600,54.55)
(700,41.7)
(800,43.15)
(900,41.95)
(1000,37.)

    };
\

\legend{}

\end{axis}

\begin{axis}[
    at={(ax1.south east)},
    xshift=1.5cm,
    title={AUCROC},
    xlabel={dimension},
    ylabel={},
    ymajorgrids=true,
    grid style=dashed,
    legend style={font=\small,legend columns=1},
]
\addlegendentry{Mahalanobis}
\addplot[smooth,color=blue,mark=square*,solid]  
    coordinates {
    (1,100)
(100,99.2658)
(200,95.9974)
(300,91.66855)
(400,90.014175)
(500,84.138025)
(600,79.261725)
(700,75.549525)
(800,71.38555)
(900,65.052225)
(1000,54.62775)

    };
\addlegendentry{LOF}
\addplot[smooth,color=red,mark=triangle*,solid] 
    coordinates {
    (1,100)
(100,99.2898)
(200,97.329775)
(300,94.9931)
(400,93.25665)
(500,90.054225)
(600,89.238075)
(700,85.8496)
(800,85.876825)
(900,84.03455)
(1000,82.720025)
    };
        
\end{axis}
\end{tikzpicture}

\caption{
Comparison of the TNR at TPR 95 \% and the AUC for Mahalonobis and LOF on simulation data. By increasing the complexity of the problem by expanding the input data dimension, the LOF method is much more stable and achieves better results. The dimension above 1,000 is common in the lasts layers of CNNs.}
\label{plot:LOFvsMahalSim}
\end{figure}


%% file: sections/analysis.tex
\section{Experimental results}

In this section, we demonstrate the performance of the proposed OoD detection using popular CNN architectures ResNet-101 and EfficientNet-B3 on several image recognition datasets. The experiments are reproducible using the code on GitHub. 

\subsection{Experiment organization}
\label{section:experiment_organization}
We decided to use low- and high-resolution datasets in our experiments. That allows comprehensive evaluation of the methods. For the low-resolution problem, we decided to use the CIFAR-10 trained by us on the ResNet-101 model. We achieved 94.79\% accuracy. Next, we chose two types of out-of-distribution datasets - the SVHN and the CIFAR-100. For the high-resolution problem, we decided to use ImageNet as known data, working with two architectures, the ResNet-101 and EfficientNet-B3. We used publicly available pre-trained versions for these models. We used three out-of-distribution datasets: ImageNet-2010, the Places365, and ImageNet-O \cite{hendrycks2019natural}. The ImageNet-2010 dataset includes images from ILSVRC 2010, with classes unknown in the ImageNet dataset.

We used the training partitions of the in-distribution datasets to build the OoD models. To evaluate the OoD methods, we used the testing partitions of the in-distribution datasets and the given out-of-distribution dataset, with a 1:1 proportion of known and unknown samples. To simplify the calculations for the ImageNet variants, in most experiments we used only the 50 first classes from the ImageNet as known data. We note that the model was trained on all 1000 classes.

We use the standard metrics for evaluation of OoD performance: TNR at TPR 95\%, AUROC, detection accuracy, and AUPR. 

Since on most benchmarks, the methods compared achieve high performance, we consider these OoD problems simple. We note that the most challenging problems reported in this study are the CIFAR-10 (as the known data) vs the CIFAR-100 (as the unknown data) and the ImageNet vs. the ImageNet-2010. 


In table \ref{tab:main_results} we present a comparison of the original Mahalanobis distance-based method by \citet{lee2018simple} with our LOF-based methods. Our approaches do not use the input pre-processing technique and use the LOF-based method for the computing confidence score - instead of the parametric approach based on Mahalanobis distance. In the study, we evaluated the simplified method, based only on the last layer of the CNN. This approach is hyperparameter-free and does not use OoD examples for model training. In table \ref{tab:main_results_simple}, we presented results for this simplified version, with results for the ImageNet dataset.

The important part of all OoD detection methods is the features extraction technique. The standard approach is the Global Average Pooling (gap). We believe that the impact of the feature generation method on OoD detection is worth in-depth testing. We used the classic gap method for the results presented in table \ref{tab:main_results}, and we tested other variants for our simplified $_{s}$LOF and $_{s}$LOF\_D methods presented in table \ref{tab:main_results_simple}. There are well-known methods like Global Maximum Pooling (gap), Cross-dimensional weighting (CroW)\cite{kalantidis2016cross}, Deep Local Feature (DeLF)\cite{noh2017large}, Selective Convolutional Descriptor Aggregation (SCDA)\cite{wei2017selective} or Generalized Mean Pooling (GeM)\cite{radenovic2018fine}. We do not recommend any specific method - the effectiveness of the methods in the context of OoD detection is the field for further research.

\subsection{Results and discussion}
\label{section:sanalysis}
In table \ref{tab:main_results}, we summarize the effectiveness of the proposed methods in comparison with the Mahanalobis distance-based method proposed by \citet{lee2018simple}.

\input{sections/table_main_results}

\input{sections/table_simple}

We observe that on difficult OoD problems (ie. CIFAR-10 as in-distribution vs CIFAR-100 as OoD; ImageNet as in-distribution vs ImageNet-2010 as OoD -- difficult problems for ResNet-101), our LoF-based procedure outperforms the Mahalanobis distance-based method. It is worth noting that this is done with fewer hyperparameters: the Mahalanobis method needs finetuning the input perturbation hyperparameter and the weights ($\alpha$ weights in Figure \ref{plot:idea}) used to aggregate the per-layer confidence scores (both require OoD samples for parameter finetuning). Our method uses only the latter hyperparameter (the $\alpha$ weights). We also observe that the optimal value of the input perturbation is study/data-dependent, in table \ref{tab:main_results} we report results with the perturbation parameter fine-tuning for this particular study. Our method is more robust in this respect.

Secondly, we observe that for the OoD benchmarks which seem easier, ie. the AUROC obtained by the methods ~ 98\% - 99\%, (these benchmarks include: CIFAR-10 vs SVHN, ImageNet vs ImageNet-O or Places365, ImageNet vs ImageNet-2010 (for EfficientNet-B3)), our method achieves similar results, again with no input perturbation hyperparameter, fine-tuned for a specific study in the Mahalanobis method.

Finally, we found that for these simple OoD benchmarks, the LOF-based score allows us to successfully detect OoD inputs using only the last layer of the network. In Table \ref{tab:main_results_simple}, we report these results. We note that this simplified LOF-based OoD detector requires no hyperparameters, and hence it does not require OoD samples for hyperparameter tuning. The method by \citet{lee2018simple} uses OoD samples to fine-tune both the input perturbation and the $\alpha$ weights hyperparameters.
 
We observe that the effectiveness of the LOF-based OoD detector using only the last layer method depends on the feature extraction method. For the ResNet-101 model and the $_{s}$LOF, we used the Euclidean distance and CroW \cite{kalantidis2016cross} method, and for the $_{s}$LOF\_D cosine distance and Global Maximum Pooling (gmp) strategy. For the EfficientNet-B3 model, we used the exact distances strategy. Moreover, we used Global Average Pooling(gap) for the $_{s}$LOF and concatenated features gap and gmp for $_{s}$LOF\_D. In-depth analysis of the effect of the feature generation method which aims to both in-distribution data discrimination and in-distribution vs out-of-distribution discrimination remains as the area of further research.

%% file: sections/table_main_results.tex
\begin{table}
  \caption{The comparison of \citet{lee2018simple} and our methods. In contrast to \cite{lee2018simple}, we do not use the input preprocessing and use LOF or LOF\_D for computing confidence scores. For the two most challenging problems, the CIFAR-10 vs. the CIFAR-100 and the ImageNet vs. the ImageNet-2010 for the ResNet-101, our approach outperforms \cite{lee2018simple}. For the rest simpler tasks, we achieved similar results.}
  \label{tab:main_results}
  \centering
  \begin{tabular}{lllcccc}
    \toprule
            \begin{tabular}{c}In-dist \\ (Model) \end{tabular} & OOD & 
            Method&
            \begin{tabular}{c}TNR at\\ TPR 95\%\end{tabular} &
            AUROC &
            \begin{tabular}{c}DTACC\end{tabular} &
            AUPR \\
    \midrule
    \multirow{6}{*}{\begin{tabular}{l}CIFAR-10 \\ (ResNet-101)\end{tabular}}   & \multirow{3}{*}{{CIFAR-100}} & Mahal. &53.77&	90.47&	83.18&	90.12 \\
    & & LOF & 63.91	&92.41&	85.24&	92.31 \\
    & & LOF\_D &\textbf{65.92}&	\textbf{93.16}&	\textbf{85.89}&	\textbf{93.11} \\
    \cmidrule{2-7}
    
    & \multirow{3}{*}{SVHN} &  Mahal. &99.02&	99.46&	97.14&	99.40 \\
    & & LOF &\textbf{99.97}&\textbf{99.95}&\textbf{99.25}&\textbf{99.94}\\
    & & LOF\_D &99.74	&99.9	&98.77	&99.9 \\

    \midrule
    \multirow{9}{*}{\begin{tabular}{l}ImageNet \\ (ResNet-101)\end{tabular}} 
    & \multirow{3}{*}{\begin{tabular}{l}{ImageNet}\\{-2010}\end{tabular}} & Mahal. & 71.60&	93.70&	\textbf{87.75}&	93.24\\
    & & LOF &\textbf{74.50}&	\textbf{94.23}&	87.28&	\textbf{94.16} \\
    & & LOF\_D &66.15&	92.31&	84.45&	92.23\\
    \cmidrule{2-7}
    & \multirow{3}{*}{\begin{tabular}{l}ImageNet\\-O\end{tabular}} &  Mahal. &\textbf{99.93}&99.88&	98.73&	99.85 \\
    & & LOF &99.67&	\textbf{99.91}&	\textbf{98.77}&	\textbf{99.88} \\
    & & LOF\_D &\textbf{99.33}&	99.77&	98.40&	99.74\\
    \cmidrule{2-7}
    & \multirow{3}{*}{Places365} & Mahal. &\textbf{99.40}	&99.64	&\textbf{97.75}&	99.62\\
            
    & & LOF & 98.65	&\textbf{99.69}&	97.32&	\textbf{99.68}\\
    & & LOF\_D &98.90&	99.66&	97.45&99.65 \\
    \midrule
     \multirow{9}{*}{\begin{tabular}{l}ImageNet \\ (EfficientNet\\-B3)\end{tabular}} 
    & \multirow{3}{*}{\begin{tabular}{l}{ImageNet}\\-2010\end{tabular}} &  Mahal. &\textbf{91.75}&	\textbf{98.11}&	\textbf{93.65}&	97.98 \\
    & & LOF &91.05&	98.08&	93.35&	\textbf{98.03}\\
    & & LOF\_D &77.75&	95.44&	88.75&	95.48\\
    \cmidrule{2-7}
    & \multirow{3}{*}{\begin{tabular}{l}ImageNet\\-O\end{tabular}} & Mahal. &99.93&99.97&	99.27&	99.93\\
    & & LOF &\textbf{99.94}&	\textbf{99.98}&	\textbf{99.63}&	\textbf{99.95}\\
    & & LOF\_D &99.25&	99.87&	98.62&	99.85 \\
    \cmidrule{2-7}
    & \multirow{3}{*}{Places365} &  Mahal. &\textbf{100.0}&	99.97&	99.48&	99.95\\
    & & LOF & 99.95&\textbf{99.99}&	\textbf{99.55}&	\textbf{99.96}\\
    & & LOF\_D &99.85&	99.89&	99.03&	99.87\\

    \bottomrule
  \end{tabular}
\end{table}

%% file: sections/table_simple.tex
\begin{table}
  \caption{The comparison of \citet{lee2018simple} and our simplified methods. In contrast to \cite{lee2018simple}, we do not use the input preprocessing, use LOF or LOF\_D for computing confidence score, but this time using only global features, our methods are hyperparameters free and do not use any OoD data. The features extraction method can be changed from the classic gap approach. We achieve comparable results for most tasks. For the ImageNet vs. the ImageNet-2010 with the ResNet-101, our methods perform better results, and for the EfficientNet-B3 worst.}
  \label{tab:main_results_simple}
  \centering
  \begin{tabular}{lllcccc}
    \toprule
            \begin{tabular}{c}In-dist \\ (Model) \end{tabular} & OOD & Method &
            \begin{tabular}{c}TNR at\\ TPR 95\%\end{tabular} &
            AUROC &
            ATACC &
            AUPR \\    

    \midrule
    \multirow{9}{*}{\begin{tabular}{l}ImageNet \\ (ResNet-101)\end{tabular}} 
    & \multirow{3}{*}{\begin{tabular}{l}ImageNet\\-2010\end{tabular}} 
    & Mahal.& 71.60 & 93.70 & 87.75 & 93.24 \\
    & & $_{s}$LOF & 56.36 & 88.40 & 80.82 & 87.57 \\
    & & $_{s}$LOF\_D & 85.40 & 92.35 & 90.32 & 90.91 \\
    \cmidrule{2-7}
    & \multirow{3}{*}{\begin{tabular}{l}ImageNet\\-O\end{tabular}} 
    & Mahal.& 99.93 & 99.88 & 98.73 & 99.85 \\
    & & $_{s}$LOF & 92.40 & 98.46 & 93.91 & 98.45 \\
    & & $_{s}$LOF\_D  & 99.21 & 99.75 & 96.87 & 96.84 \\
    \cmidrule{2-7}
    & \multirow{3}{*}{\begin{tabular}{l}Places365\end{tabular}} 
    & Mahal.& 99.40 & 99.64 & 99.75 & 99.62 \\
    & & $_{s}$LOF & 91.40 & 97.87 & 93.18 & 97.88 \\
    & & $_{s}$LOF\_D  & 99.68 & 97.64 & 97.56 & 99.58 \\
    
    \midrule
    
    \multirow{9}{*}{\begin{tabular}{l}ImageNet \\ (EfficientNet-B3)\end{tabular}} 
    & \multirow{3}{*}{\begin{tabular}{l}ImageNet\\-2010\end{tabular}} 
    & Mahal.& 91.75 & 98.11 & 93.65 & 97.98 \\
    & & $_{s}$LOF & 69.27 & 89.24 & 83.94 & 86.07 \\
    & & $_{s}$LOF\_D  & 84.61 & 87.64 & 90.07 & 80.95 \\
    \cmidrule{2-7}
    & \multirow{3}{*}{\begin{tabular}{l}ImageNet\\-O\end{tabular}} 
    & Mahal.& 99.93 & 99.97 & 99.27 & 99.93 \\
    & & $_{s}$LOF & 97.82 & 99.46 & 97.27 & 99.44 \\
    & & $_{s}$LOF\_D  & 99.06 & 99.17 & 97.32 & 98.59 \\
    \cmidrule{2-7}
    & \multirow{3}{*}{\begin{tabular}{l}Places365\end{tabular}} 
    & Mahal.& 100.0 & 99.97 & 99.48 & 99.95 \\
    & & $_{s}$LOF & 98.20 & 99.51 & 96.90 & 99.48 \\
    & & $_{s}$LOF\_D  & 99.64 & 98.31 & 98.02 & 97.44 \\
    \bottomrule

  \end{tabular}
\end{table}



%% file: sections/conclusion.tex
\section{Conclusion}
\label{section:conclusion}
We proposed a simple method for detecting out-of-distribution inputs in a DNN. The method uses standard representations learned by modern neural network architectures and does not require network retraining. Our method uses LOF-based scores to quantify predictive uncertainty for test samples, in place of Mahalanobis distance-based confidence scores used in SOTA OoD detectors which rely on the standard representation. In several feasibility studies, we showed that this modification leads to improved performance with regard to Manalanobis-based OoD detectors, or realizes similar performance in a simpler way. Our method uses fewer hyperparameters (on none in a simplified version): it does not rely on the input perturbation hyperparameter used by current methods such as \citep{liang2017enhancing} or \citep{lee2018simple}. We showed that for simple OoD problems, a simplified LOF-based quantification of outlierness of inputs using only the last layer features can realize similar performance as complex, hyperparameter-dependent Mahalanobis distance-based SOTA methods such as \citep{lee2018simple}. We note that this simplified approach is entirely hyperparameter-free, and as such does not require OoD samples for hyperparameter tuning, contrary to Mahalanobis SOTA approaches.   
Further improvement of our method may come from in-depth analysis of the effect of feature generation procedures (such as gmp, gap, CroW, DeLF, SCDA, GeM, etc.) at different layers of the network (of at the last layer when using the simplified method). Since our method performs OoD detection using standard representations learned to maximize in-distribution discrimination, an interesting question remains which methods of feature extraction are most effective if we want to achieve successful in-distribution discrimination as well as in- vs out-of-distribution discrimination. 

\textbf{Limitations.} We are aware of the following limitations of the proposed method.

Although the effectiveness of OoD detection using LOF-based scores depends on the method of feature generation from the subsequent layer of the network, we do not provide a systematic analysis of this effect. We believe the gap (global average pooling) method yields reasonable performance, based on results reported in Table \ref{tab:main_results}. However, OoD detection based on the last layer features seems more sensitive to the choice of feature generation, and the choice may be data-dependent. This limitation motivates further research.

We do not report variability of results in tables \ref{tab:main_results} and \ref{tab:main_results_simple} with respect to random train/validation/test partitioning of the data. We work with relatively large, class-balanced datasets, with ca. 1000-5000 points per class, hence we believe the variability is not significant (we have some point observations of e.g., TNR changing ca. 0.1-1 percentage points with different partitions).